\documentclass[a4paper, 10pt, conference]{ieeeconf}      

\IEEEoverridecommandlockouts                              

\usepackage{xcolor}
\usepackage{soul}
\usepackage[utf8]{inputenc}
\usepackage[small]{caption}
\usepackage{booktabs}
\usepackage{graphics} 
\usepackage{tabularx}
\usepackage{amsmath} 
\usepackage{amssymb}  
\usepackage{mathrsfs}
\usepackage{subcaption}
\usepackage{tikz}
\usetikzlibrary{shapes,arrows, automata}
\usepackage{pgfplots}
\usepackage[per-mode=symbol]{siunitx}
\DeclareSIUnit{\mph}{mph}
\usepackage{url}
\usepackage[noadjust]{cite}
\usepackage{numprint}
\usepackage{bm}
\newcommand{\vect}[1]{\boldsymbol{\mathbf{#1}}}

\pgfplotsset{compat=newest}
\pgfplotsset{every axis/.append style={
	font=\LARGE}
}

\DeclareUnicodeCharacter{2212}{−}
\usepgfplotslibrary{groupplots,dateplot}
\usetikzlibrary{patterns,shapes.arrows}

\pgfplotsset{every axis legend/.append style={legend cell align=left}}
\def\man#1;{%
    \begin{scope}[shift={#1}]
        \fill [rounded corners=1.5] (0,0.4) -- (0,0.8) -- (0.4,0.8) -- (0.4,0.4) --
            (0.325,0.4) -- (0.325,0.7) -- (0.3,0.7) -- (0.3,0) -- (0.225,0) --
            (0.225,0.4) -- (0.175,0.4) -- (0.175,0) -- (0.1,0) -- (0.1,0.7) --
            (0.075,0.7) -- (0.075,0.4) -- cycle;
        \fill (0.2,0.9) circle (0.1);
    \end{scope}}
    
\newcommand{\argmax}{\operatornamewithlimits{arg\,max}}

\usetikzlibrary{arrows.meta, calc, shapes}
\tikzset{%
  >={Latex[width=2mm,length=2mm]},
            base/.style = {rectangle, rounded corners, draw=black,
                           minimum width=1cm, minimum height=1cm,
                           text centered, font=\sffamily},
            simulator/.style = {base, fill=green!30, minimum width=4cm},
            solver/.style = {base, fill=red!30},
            reward/.style = {base, minimum height=1.5cm},
            module/.style = {base, minimum width=2.5cm, minimum height=1.5cm, fill=blue!30},
            module2/.style = {base, minimum width=2.5cm, minimum height=1.5cm, fill=white},
            network/.style = {base, minimum width=2.5cm, minimum height=1.5cm, fill=white},
            state/.style = {base, minimum width=0.5cm, minimum height=1.0cm, fill=white},
            io/.style = {base, minimum width=0.5cm, minimum height=1.0cm, fill=white},
}
\pgfdeclarelayer{background}
\pgfdeclarelayer{foreground}
\pgfsetlayers{background,main,foreground}

\newif\ifcuboidshade
\newif\ifcuboidemphedge

\tikzset{
  cuboid/.is family,
  cuboid,
  shiftx/.initial=0,
  shifty/.initial=0,
  dimx/.initial=3,
  dimy/.initial=3,
  dimz/.initial=3,
  scale/.initial=1,
  densityx/.initial=1,
  densityy/.initial=1,
  densityz/.initial=1,
  rotation/.initial=0,
  anglex/.initial=0,
  angley/.initial=90,
  anglez/.initial=225,
  scalex/.initial=1,
  scaley/.initial=1,
  scalez/.initial=0.5,
  front/.style={draw=black,fill=white},
  top/.style={draw=black,fill=white},
  right/.style={draw=black,fill=white},
  shade/.is if=cuboidshade,
  shadecolordark/.initial=black,
  shadecolorlight/.initial=white,
  shadeopacity/.initial=0.15,
  shadesamples/.initial=16,
  emphedge/.is if=cuboidemphedge,
  emphstyle/.style={thick},
}

\makeatother

\makeatletter
\let\NAT@parse\undefined
\makeatother
\usepackage{hyperref} 
\usepackage{cleveref}

\usepackage[keeplastbox]{flushend}

\title{\Large \bf Adaptive Stress Testing without Domain Heuristics using Go-Explore}
\author{Mark Koren$^{1}$ and Mykel J. Kochenderfer$^{1}$
\thanks{$^{1}$Mark Koren and Mykel J. Kochenderfer are with Aeronautics and Astronautics, Stanford University, Stanford, CA 94305, USA
        {\tt\small \{mkoren, mykel\}@stanford.edu}}
}

\begin{document}

\maketitle
\thispagestyle{empty}
\pagestyle{empty}

\begin{abstract}  

Recently, reinforcement learning (RL) has been used as a tool for finding failures in autonomous systems.
During execution, the RL agents often rely on some domain-specific heuristic reward to guide them towards finding failures, but constructing such a heuristic may be difficult or infeasible.
Without a heuristic, the agent may only receive rewards at the time of failure, or even rewards that guide it away from failures. 
For example, some approaches give rewards for taking more likely actions, in order to to find more likely failures.
However, the agent may then learn to only take likely actions, and may not be able to find a failure at all.
Consequently, the problem becomes a hard-exploration problem, where rewards do not aid exploration.
A new algorithm, go-explore (GE), has recently set new records on benchmarks from the hard-exploration field.
We apply GE to adaptive stress testing (AST), one example of an RL-based falsification approach that provides a way to search for the most likely failure scenario.
We simulate a scenario where an autonomous vehicle drives while a pedestrian is crossing the road.
We demonstrate that GE is able to find failures without domain-specific heuristics, such as the distance between the car and the pedestrian, on scenarios that other RL techniques are unable to solve.
Furthermore, inspired by the robustification phase of GE, we demonstrate that the backwards algorithm (BA) improves the failures found by other RL techniques.
\end{abstract}

\section{Introduction}\label{sec:intro}
Safety validation remains a challenge in the development of autonomous vehicles~\cite{Koopman2018,Kalra2016}. 
A promising approach is to treat validation as a reinforcement learning (RL) falsification task, in which the goal is to find scenarios that lead to a system failure using RL.
For example, Akazaki et al. apply deep reinforcement learning (DRL) methods to perform falsification of signal temporal logic properties of cyber-physical systems~\cite{akazaki2018falsification}.
Zhang et al. use Monte Carlo tree search (MCTS) in conjunction with hill-climbing to perform optimization-based falsification on hybrid systems~\cite{akazaki2018falsification}.
Delmas et al. use MCTS for property falsification on hybrid flight controllers~\cite{delmas2019evaluation}. 
Adaptive stress testing (AST)~\cite{lee2015adaptive}, an RL-based method for finding the most likely failure, and is the falsification approach used in this paper.

\begin{figure}[t]
    \setlength\belowcaptionskip{-0.75\baselineskip}	
	\centering
    \scalebox{0.8}{\begin{tikzpicture}[node distance=1.5cm,
    every node/.style={fill=white, font=\sffamily, text centered}, align=center]
	\node (sim)             [simulator]              {Simulator $\mathcal{S}$};
    \node (solver)          [solver, above of = sim, xshift = 0.8cm]              {Solver};
    \node (reward)          [reward, above of = sim, xshift = 5cm]              {Reward\\Function};  
  \draw[->]					(solver.west) -| node[text width=1cm, xshift = 0mm, yshift = 5mm, text centered, align=center] {Environment\\Actions} ($ (sim.north) - (15mm, 0) $);
  \draw[->] 	($(sim.east) + (0mm,2mm)$) -| node[text width=1.6cm, xshift = -17mm, yshift = 4mm] {Likelihood} ($ (reward.south) + (-2mm, 0mm) $);
  \draw[->] 	($(sim.east) + (0mm,-2mm)$) -| node[text width=1cm, xshift = -20mm, yshift = -4mm] {Event} ($ (reward.south) + (2mm, 0mm) $);
  \draw[->]		(reward.west) -- ++(0mm,0) -- node[text width=1cm, xshift = 0mm, yshift = 3mm] {Reward}(solver.east);
\end{tikzpicture}}
    \caption{The AST methodology. The simulator is treated as a black box. The solver optimizes to maximize a reward based on action likelihood and whether an event has occurred.}
	\label{fig:ASTStruct} 
	\vspace{-1mm}
\end{figure}
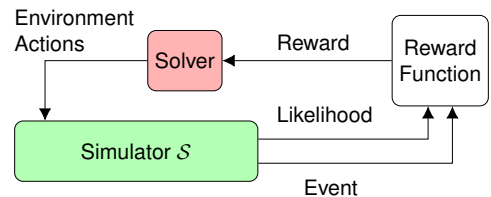

In AST, we treat both the system under test and the simulation as a black-box environment that is controlled by actions from the AST solver, which acts as the agent in our Markov decision process (MDP)~\cite{Koren}.
Therefore, falsification becomes a sequential decision making problem, which reinforcement learning is well-suited to solve.
The formulation provides two distinct advantages: 1) we can use RL to efficiently search the space of possible actions for failures and 2) we receive an approximation of the most likely error, which allows designers to characterize how their system is most vulnerable to failure.
The process is shown in \cref{fig:ASTStruct}, where the AST solver takes actions and then receives rewards from the simulation.
The reward can be conceptually decomposed into two parts: 1) A large penalty applied when the trajectory does not end in failure and 2) a cost at each timestep proportional to the negative log-likelihood of actions taken, such that more likely actions are penalized less. 
The scale of the rewards is such that the AST agent should prioritize unlikely failures over likely non-failures, but also prefer more likely failures to less likely failures.
Consequently, maximizing reward leads to the most likely failure.

A difficulty arises from the fact that the failure-finding cost is only applied at the end of trajectories. 
Even worse, the likelihood component applied at each timestep does not actually inform the AST agent on how to find a failure. 
A mature system would only fail for a series of unlikely actions, so the per-step reward may actually be leading the agent away from failures.
One approach is to add a heuristic reward (for example, the miss distance between a pedestrian and a car) to guide the agent towards failures~\cite{Koren}, but creating a heuristic may be difficult or infeasible. 

In reinforcement learning, a problem where reward signals are rare is known as a sparse rewards problem or a hard-exploration problem~\cite{bellemare2016unifying}.
Solving hard-exploration problems is an active subset of RL research, with one of the most famous canonical baselines being the Atari game Montezuma's Revenge~\cite{burda2018exploration, mnih2015human, lipovetzky2015classical}.
Montezuma's Revenge requires completing multiple sub-tasks that do not give reward in order to advance to the next level, and the horizon is sufficiently long such that the agent may never randomly discover the right sequence. 
Recently, the newly released go-explore (GE) algorithm was able to set new records on Montezuma's Revenge~\cite{ecoffet2019go}.

The go-explore algorithm has two phases.
Phase 1 consists of a biased random tree search that has shown state-of-the-art performance on hard-exploration problems with long horizons, like Montezuma's Revenge.
While validation tasks may not have trajectories nearly as long as Montezuma's Revenge, having horizons of tens or even hundreds of steps is possible, especially when working with high-fidelity simulators.
Phase 2 consists of ``robustifying'' the results by training a neural network policy using the output of  phase 1 as an expert demonstration.
While a falsification trajectory does not need to be robust to noise in a traditional sense, phase 2 does allow us to use the strengths of deep learning to improve our results from phase 1 while giving us more coverage of a scenario.

One of the advantages of the AST approach is that independent advancements in RL, a much larger field of research, can be applied to the solvers to improve the validation of autonomous systems~\cite{koren2019efficient, corso2019adaptive}.
We demonstrate that principle by modifying GE to work on an autonomous vehicle validation task, and comparing it to the results of two previous RL solvers that we have presented: 1) a Monte Carlo tree search (MCTS) solver, and 2) a deep reinforcement learning (DRL) solver.
This paper contributes the following:
\begin{itemize}
    \item We present a modification of GE better suited for general validation tasks.
    \item We demonstrate that using GE allows us to find failures on hard-exploration scenarios for which MCTS and DRL could not find failures.
    \item We demonstrate that applying the robustification phase to our previous solvers allows us to improve our validation results on scenarios that do not involve hard-exploration.
\end{itemize}
These contributions advance the ability of AST to find failures as well as to maximize their likelihood.

This work is organized as follows. 
\Cref{sec:background} outlines the concepts of AST and RL.
\Cref{sec:methodology} explains the modifications made to GE for validation.
\Cref{sec:experiment} provides the implementation details of our experiment.
\Cref{sec:Results} demonstrates GE on longer-horizon validation tasks and shows the benefits of a robustification phase. 

\section{Adaptive Stress Testing}\label{sec:background}

\subsection{Markov Decision Process}\label{sec:bg_markov}
AST frames the problem of finding a failure in simulation as a Markov decision process (MDP).
At each timestep, an agent takes action $a$ from state $s$.
The agent may receive a reward based on the reward function $R(s,a)$~\cite{DMU}. 
The agent then transitions to the next state $s'$ based on the transition function $P(s'\mid s,a)$. 
The reward and transition depend only on the current state-action pair $(s,a)$, not the history of previous actions or states, an independence assumption known as the Markov assumption. 
Neither the reward nor transition functions need to be deterministic.

Agents behave according to a policy $\pi(s)$, which specifies which action to take from each state. 
A policy can be stochastic or deterministic. 
The optimal policy is one that maximizes the value function, which can be found for a policy $\pi$ recursively:
\begin{equation}
V^{\pi} \left(s\right) = R\big(s, \pi\left(s\right)\big) + \gamma  \sum_{s'}P\big(s'\mid s,\pi \left(s\right)\big)V^{\pi}\left(s'\right)
\end{equation}
where $\gamma$ is the discount factor that controls the weight of future rewards.
For smaller MDPs, the value function can be solved exactly. 
Larger MDPs may be solved approximately using reinforcement learning techniques.

\subsection{Formulation}\label{sec:bg_ast}
The motivational problem for AST is finding the most likely trajectory that ends in a target set:
\begin{equation}
\label{eq:ast_optimzation}
\begin{aligned}
& \underset{a_0, \ldots, a_t}{\text{maximize}}
& & P(s_0, a_0, \ldots,s_t, a_t) \\
& \text{subject to}
& & s_t \in E
\end{aligned}
\end{equation}
where $P(s_0, a_0, \ldots,s_t, a_t)$ is the probability of a trajectory in simulator $\mathcal{S}$ and where, by the Markov assumption, $s_t$ is only a function of $a_t$ and  $s_{t-1}$. The set $E$ is generally defined to contain failure events. For example, for an autonomous vehicle $E$ might include collision states and near-misses. 

We use reinforcement learning (RL) to approximately solve \cref{eq:ast_optimzation}.
A trajectory that maximizes \cref{eq:ast_optimzation} also maximizes $\sum_{t=0}^{T} R(s_t, a_t)$ where 
\begin{equation}
 \label{eq:base_reward}
R\left(s_t, a_t\right) = \left\{
        \begin{array}{ll}
            0, &  s_t \in E \\
            -\infty, &  s_t \notin E, t\geq T \\
            -\log \left( P(a_t \mid s_t)\right),   &  s_t \notin E, t < T 
        \end{array}
    \right.
\end{equation}
and $T$ is the horizon~\cite{koren2019adaptive}.

AST solvers can treat the simulation and system under test as black-box components, however the following access functions must be provided by the simulator:
\begin{itemize}
\item \textsc{Initialize}$(\mathcal{S}, s_0)$: Resets $\mathcal{S}$ to a given initial state $s_0$.
\item \textsc{Step}$(\mathcal{S}, E, a)$: Steps the simulation in time by drawing the next state $s'$ after taking action $a$. The function returns the probability of the action taken and an indicator whether $s'$ is in $E$ or not.
\item \textsc{IsTerminal}$(\mathcal{S}, E)$: Returns true if the current state of the simulation is in $E$, or if the horizon of the simulation $T$ has been reached. 
\end{itemize}
We have previously presented a DRL solver as well as an MCTS solver~\cite{Koren}. In this paper, we also present a solver based on the go-explore algorithm. 

\subsection{Deep Reinforcement Learning}\label{sec:bg_drl}
In deep reinforcement learning (DRL), a policy is represented by a neural network parameterized by $\theta$~\cite{goodfellow2016deep}.
While a feed-forward neural network maps an input to an output, we use a recurrent neural network (RNN), which maps an input and a hidden state from the previous timestep to an output and an updated hidden state.
An RNN is naturally suited to sequential data due to the hidden state, which is a learned latent representation of the current state. 
RNNs suffer from exploding or vanishing gradients, a problem addressed by variations such as long-short term memory (LSTM)~\cite{Hochreiter1997} or gated recurrent unit (GRU)~\cite{bahdanau2014neural} networks.

There are many different algorithms for optimizing a neural network, but proximal policy optimization (PPO)~\cite{schulman2017proximal} is one of the most popular.
PPO is a policy-gradient method that updates the network parameters to minimize the cost function.
Improvement in a policy, compared to the old policy, is measured by an advantage function.
There are a variety of ways to estimate an advantage function, such as generalized advantage estimation (GAE)~\cite{Schulman2015}, which allows optimization over batches of trajectories. 
However, variance in the advantage estimate can lead to poor performance if the policy changes too much in a single step.
To prevent this, PPO can limit the step size in two ways: 1) by incorporating a penalty proportional to the KL-divergence between the new and old policy or 2) by clipping the estimated advantage function when the new and old policies are too different. 

\subsection{Monte Carlo Tree Search}\label{sec:bg_mcts}
Monte Carlo tree search (MCTS)~\cite{MCTSUCT} is one of the best-known planning algorithms, and has a long history of performing well on large MDPs~\cite{MCTS_GO}.
MCTS is an anytime online sampling-based method that builds a tree where the nodes represent states and actions.
While executing from states in the tree, MCTS chooses the action that maximizes   
\begin{equation}
\label{eq:UCB}
\argmax_a Q(s,a)+c\sqrt{\frac{\log(N(s))}{N(s,a)}}
\end{equation}
where $Q(s,a)$ is the expected return associated with a state-action pair, $N(s)$ and $N(s,a)$ are the number of times a state and a state-action pair have been visited, respectively, and $c$ is a parameter that controls exploration.
The value $Q(s,a)$ is updated by executing rollouts to a specified depth and then returning the reward, which is then back-propagated up the tree. 
This paper uses a variation of MCTS with double progressive widening (DPW)~\cite{MCTSDPW} to limit the branching of the tree, which can improve performance on problems with large or continuous action spaces.

\subsection{Go-Explore}\label{sec:bg_ge}
The go-explore (GE) algorithm is designed for hard-exploration problems~\cite{ecoffet2019go}.
The algorithm consists of an exploration-until-solved phase and a robustification phase.

\subsubsection{Phase 1 \textemdash Explore until solved}\label{sec:bg_ge_phase1}
Phase 1 is essentially a biased random tree search that takes advantage of determinism for exploration. 
During phase 1, a pool of ``cells'' is maintained, where a cell is a data-structure indexed by a compressed mapping of the agent's state.
During rollouts, any ``unseen'' cell is added to the pool. If a cell has already been seen before, we compare the old version and the newly visited version.
If the new cell has a better score, or has a shorter trajectory before being reached, then we replace the old cell with the new cell.
Every rollout starts by randomly sampling a cell from the cell pool, which serves as the starting condition. 
Fundamentally, the agent deterministically follows the same trajectory back to the sampled cell, and then explores from there. 
The authors present multiple strategies for sampling cells from the cell pool to achieve better results.
The exploration process continues for a set number of iterations or until a solution is found.

\subsubsection{Phase 2 \textemdash Robustification}\label{sec:bg_ge_phase2}
Phase 2 uses the best result from phase 1 to train a robust policy.
The result from phase 1 serves as our expert demonstration in an imitation learning task, also known as learning from demonstration (LfD).
While other imitation learning algorithms could be used, the authors use the backwards algorithm (BA)~\cite{salimans2018learning}, because of its potential to improve upon the expert demonstration.
Under BA, a policy first starts training by executing rollouts from the last step of the expert trajectory.
Any standard deep-learning optimization technique, such as PPO, can be used for training.
The policy trains from this step of the trajectory until it does as well or better than the expert's score.
Then the starting point of training rollouts is moved one step back in the expert trajectory, to the second-to-last step, and training repeats.
The process continues training and moving the starting point back until the agent is able to do as well or better than the expert on rollouts from the start of the expert trajectory. 
The deep-learning agent is able to efficiently learn to overcome stochastic deviations from the expert trajectory, and may discover and exploit even better strategies, if the expert trajectory is not optimal. 

\section{Methodology}\label{sec:methodology}
This paper applies GE to a validation task without the use of a domain-specific heuristic reward.
We show that phase 1 of the algorithm is able to find failures without the use of heuristics on problem instantiations that other solvers fail on.
Furthermore, we are able to use phase 2 of the algorithm to improve the results of all RL solvers that we have previously published.
This section explains our modifications to both phases of the algorithm to apply GE to falsification. 

\subsection{Phase 1}\label{sec:method_phase1}

In GE, a cell is indexed by a compressed representation of the simulator state, which is also used to deterministically reset the simulation to a specific cell. 
AST is designed to be able to treat the simulator as a black-box (although benefits can be obtained if white-box simulation information is available).
To preserve the black-box assumption, we index cells by hashing a concatenation of the current step number $t$ with the discretized action $\widetilde{a_t}$, so the index is $idx = \text{hash}\left(\left[t, \widetilde{a_t} \right] \right)$.
Therefore, similar actions at the same step of the simulation will be treated as the same cell, preserving the black-box assumption by eliminating the dependence on simulation state.
Furthermore, by tracking the history of actions taken to arrive at each cell, we are able to deterministically reset to a cell without the simulation state.
Instead, the agent deterministically repeats the actions in the cell's history.

A key component of phase 1 of GE is the cell selection heuristic.
Cells are assigned a fitness score, which is then normalized across all cells. 
Every rollout starts from a cell selected at random from the pool with probability equal to the normalized fitness score of each cell. 
The fitness score is partially made up of ``count subscores'' for three attributes that represent how often a cell has been interacted with: 1) the number of times a cell has been chosen to start a rollout, 2) the number of times a cell has been visited, and 3) the number of times a cell has been chosen since a rollout from that cell has resulted in the discovery of a new or improved cell. 
For each of these three attributes, a count subscore for cell $c$ and attribute $a$ can be calculated as, 
\begin{equation}
 \label{eq:count_subscore}
    \text{CntScore}(c,a) = w_a\left(\frac{1}{v(c,a) + \epsilon_1} \right)^{p_a} + \epsilon_2
\end{equation}
where $v(c,a)$ is the value of attribute $a$ for cell $c$, and $w_a$, $p_a$, $\epsilon_1$, and $\epsilon_2$ are hyperparameters. 
The total unnormalized fitness score is then 
\begin{multline}
 \label{eq:count_score}
    \text{CellScore}(c) = \\\text{ScoreWeight}(c)\left( 1 + \sum_a \text{CntScore}(c,a) \right)
\end{multline}

When applied to Montezuma's Revenge, the authors used a $\text{ScoreWeight}$ based on what level a cell had reached, which favored cells that had progressed furthest within the game.
Progressing furthest does not have a direct analog to our validation task, so we present our own $\text{ScoreWeight}$.
Similar to MCTS, cells track an estimate of the value function. 
Anytime a new cell is added to the pool or a cell is updated, the value estimate for a specific cell $v_c$ is updated as,
\begin{equation}
 \label{eq:value_update}
    v_c \leftarrow v_c + \frac{\left(r + \gamma v_{child}^{*} \right) - v_c}{N} 
\end{equation}
where $r$ is the cell's reward, $N$ is the number of times we have seen the cell, and $v_{child}^{*}$ is the maximum value estimate of cells that are immediate children of the current cell.
When a cell is updated, it then propagates the update to its parent, and so forth up the tree.
The total unnormalized fitness score is  calculated with $\text{ScoreWeight}(c) = v_c$. Cells with high value estimates are selected more often.

\subsection{Phase 2}\label{sec:method_phase2}

Phase 2 of GE was designed to create a policy robust to stochastic noise, while potentially improving upon the expert demonstration. 
Within the context of AST, however, phase 2 acts as a hill-climbing step.
AST is designed to find the most likely failure even in high-dimensional state and action spaces. 
However, large action-spaces can make it difficult to converge reliably to a consistent result, due to lack of coverage of the action space.
While phase 1 can provide a good guess for the most likely failure trajectory, phase 2 allows us to approximately search the space of similar trajectories to find the best one.
This idea is applicable to all of the solvers we have used on AST so far, and we demonstrate this by using it to improve every failure trajectory found in this paper.
Phase 2 is most useful for the tree-search methods in particular (MCTS and GE), as they typically demonstrate better exploration in harder scenarios, but at the expense of converging to high-reward trajectories.

In the GE paper, as well as the original BA paper, a policy was trained from a specific step of the expert trajectory until it learned to do as well, or better. 
In the validation tasks we are interested in, compute may be too limited to be able to train for indefinite amounts of time. 
Instead, we modify BA to instead train for a small number of epochs at each step of the expert trajectory.
This modification allows the number of total iterations to be specified and known ahead of time, and does not prevent phase 2 from improving the expert trajectory in any of our experiments. 

\section{Experiment}\label{sec:experiment}

\begin{figure}[t]
    \setlength\belowcaptionskip{-0.75\baselineskip}	
	\centering
    \centering
    \includegraphics[width=0.70\columnwidth]{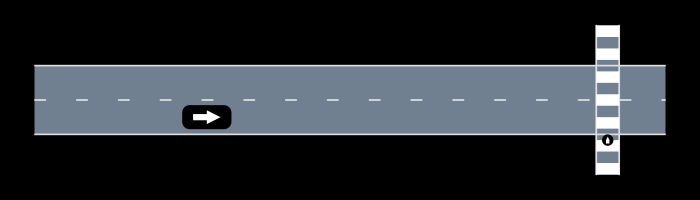}
    \caption{Layout of the crosswalk example. A car approaches a crosswalk on a neighborhood road with one lane in each direction. A pedestrian is attempting to cross the street at the crosswalk.}
	\label{fig:scenario1}
	\vspace{-1mm}
\end{figure}

\subsection{Problem Description}\label{sec:ex_setup}
The validation scenario consists of a vehicle approaching a crosswalk on a neighborhood road as a pedestrian is trying to cross, as shown in \cref{fig:scenario1}. 
The car is approaching at the speed limit of \SI{25}{\mph}(\SI{11.17}{\meter\per\second}). 
The vehicle, a modified version \cite{Koren} of the intelligent driver model (IDM)~\cite{PhysRevE621805}, has noisy observations of the position and velocity of the pedestrian. 
The AST solver controls the simulation through a six-dimensional action space, consisting of the $x$ and $y$ components for three parameters: 1) the pedestrian acceleration, 2) the noise on the pedestrian position, and 3) the noise on the pedestrian velocity.
We treat the simulation as a black-box, so the AST agent only has access to the initial conditions and the history of previous actions.
From this general set-up, we instantiate three specific scenarios, which are differentiated by the difficulty of finding a failure.
The settings changed between the scenarios include whether a reward heuristic was used (see \cref{sec:ex_reward}), the initial location of the pedestrian, as well as the rollout horizon and timestep size. 
Pedestrian and vehicle location are measured from the origin, which is located at the intersection of the center of the crosswalk and the center of the vehicle's lane.
The scenario parameters are shown in \cref{table:parameters}.
The easy scenario is designed such that the average action leads to a collision, so the maximum possible reward is known to be \num{0}.
The medium and hard scenarios require unlikely actions to be taken to force a collision. They have the same initial conditions, except the hard scenario has a timestep half the duration of the medium scenario, and therefore double the maximum path length.
The hard scenario demonstrates the effect of horizon length on exploration difficulty.  

\begin{table}[h]
    \caption{Parameters that define the easy, medium, and hard scenarios. Changing the pedestrian location makes more exploration necessary to find a collision, while changing the horizon and timestep makes exploration more complex.}
    \label{table:parameters}
    \begin{center}
        \npdecimalsign{.}
\nprounddigits{1}
\sisetup{round-mode=places,round-precision=2}
\begin{tabular}{@{}lrrr@{}}
\toprule
      Variable & Easy & Medium & Hard \\ \midrule
      {$\beta$} & {\num{1000}} & {\num{0}} & {\num{0}} \\
{$\vect{s}_{0, ped, y}$}      & {$\SI{-4}{\meter} $}  & {$\SI{-6}{\meter}$} & {$\SI{-6}{\meter}$}\\
{$T$}         & {\num{50} steps} & {\num{50} steps} & {\num{100} steps} \\
{$dt$}      & {$\SI{0.1}{\second} $}   & {$\SI{0.1}{\second} $} & {$\SI{0.05}{\second} $} \\
\bottomrule
\end{tabular}
\npnoround
    \end{center}
    \vspace{-7mm}
\end{table}

\subsection{Modified Reward Function}\label{sec:ex_reward}
We make some modifications to the theoretical reward function shown in \cref{eq:base_reward} to allow practical implementation:
\begin{equation}
 \label{eq:actual_reward}
R\left(s\right) = \left\{
        \begin{array}{ll}
            0, &  s \in E \\
            -\alpha - \beta\times\textsc{dist}\left(\vect p_v,\vect p_p\right), &  s \notin E, t\geq T \\
            - M\left(a, \mu_a, \Sigma_a\mid s\right),   &  s \notin E, t < T
        \end{array}
    \right.
\end{equation}
where $ M(a, \mu_a, \Sigma_a\mid s)$ is the Mahalanobis distance~\cite{mahalanobis1936generalised} between the action $a$ and the expected action $\mu_a$ given the covariance matrix $\Sigma_a$ in the current state $s$ and $\textsc{dist}\left(\vect p_v,\vect p_p\right)$ is the distance between the pedestrian and the vehicle at the end of the rollout.
The latter reward is the domain-specific heuristic reward that guides AST solvers by giving less penalty when the scenario ends with a pedestrian closer to the car. 
We use $\alpha = \SI{-1e5}{}$ and $\beta = \SI{-1e4}{}$ for the easy scenario, and $\beta = 0$ for the medium and hard scenarios, to disable the heuristic. 

\subsection{Solvers}\label{sec:ex_solvers}
For each experiment, the DRL, MCTS, and GE solvers were run for \num{100} iterations each with a batch size of \num{500}. 
For each solver that finds a failure, BA was then run for \num{100} iterations with a batch size of \num{5000}, with the results reported as DRL+BA, MCTS+BA, and GE+BA, respectively. 
\subsubsection{Go-Explore Phase 1}\label{sec:ex_solvers_ge}
GE was run with hyperparameters similar to those used for Montezuma's Revenge. 
For the count subscore attributes (times chosen, times chosen since improvement, and times seen), we set $w_a$ equal to \num{0.1}, \num{0}, and \num{0.3} respectively. 
All attributes share $\epsilon_1 = 0.001$, $\epsilon_2 = 0.00001$, and $p_a = 0.5$.
We always use a discount factor of \num{0.99}.
During rollouts, actions are sampled uniformly.
\subsubsection{Deep Reinforcement Learning}\label{sec:ex_solvers_drl}
The DRL solver uses a Gaussian-LSTM trained with PPO and GAE. 
The LSTM has a hidden layer size of \num{64} units, and uses peephole connections~\cite{gers2000recurrent}.
For PPO, we used both a KL penalty with factor \num{1.0} as well as a clipping range of \num{1.0}. 
GAE uses a discount of \num{0.99} and $\lambda =1.0$. 
There is no entropy coefficient. 
\subsubsection{Monte Carlo Tree Search}\label{sec:ex_solvers_mcts}
We use MCTS with DPW where rollout actions are sampled uniformly from the action space. 
The exploration constant was \num{100}. The DPW parameters are set to $k = 0.5$ and $\alpha = 0.5$. 
\subsubsection{Backwards Algorithm}\label{sec:ex_solvers_ba}
BA represents the policy with a Gaussian-LSTM, and optimizes the policy with PPO and GAE.
The hyperparameters are identical to the DRL solver.

\section{Results}\label{sec:Results}

\begin{figure}[t]
	\centering
    \centering
    \scalebox{0.72}{
\begin{tikzpicture}

\definecolor{color0}{rgb}{0.001462,0.000466,0.013866}
\definecolor{color1}{rgb}{0.578304,0.148039,0.404411}
\definecolor{color2}{rgb}{0.987622,0.64532,0.039886}

\begin{axis}[
font=\normalsize,
legend cell align={left},
legend style={fill opacity=0.8, draw opacity=1, text opacity=1, at={(0.03,0.97)}, anchor=north west, draw=white!80!black},
tick align=outside,
tick pos=left,
x grid style={white!69.0196078431373!black},
xlabel={Iterations},
xmin=-5, xmax=105,
xtick style={color=black},
y grid style={white!69.0196078431373!black},
ylabel style={align=center},
ylabel={Likelihood \\ $[$Mahalanobis Distance$]$},
restrict y to domain=-10000:0,
ymin=-700, ymax=0,
ytick style={color=black}
]
\path [draw=color0, semithick, dash pattern=on 5.55pt off 2.4pt]
(axis cs:0,-329.0490234914)
--(axis cs:100,-329.0490234914);

\path [draw=color1, semithick, dash pattern=on 5.55pt off 2.4pt]
(axis cs:0,-18.297704685)
--(axis cs:100,-18.297704685);

\path [draw=color2, semithick, dash pattern=on 5.55pt off 2.4pt]
(axis cs:0,-40.2685434172)
--(axis cs:100,-40.2685434172);

\addplot [semithick, color0]
table {%
0 -680.6057680718
1 -647.0875672097
2 -639.2264654279
3 -639.2264654279
4 -636.3201483656
5 -623.3856857326
6 -623.3856857326
7 -623.3856857326
8 -623.3856857326
9 -623.3856857326
10 -623.3856857326
11 -623.3856857326
12 -623.3856857326
13 -590.7656116329
14 -590.7656116329
15 -590.7656116329
16 -590.7656116329
17 -583.0791212443
18 -583.0791212443
19 -583.0791212443
20 -583.0791212443
21 -583.0791212443
22 -583.0791212443
23 -583.0791212443
24 -583.0791212443
25 -583.0791212443
26 -583.0791212443
27 -583.0791212443
28 -583.0791212443
29 -583.0791212443
30 -583.0791212443
31 -583.0791212443
32 -583.0791212443
33 -583.0791212443
34 -583.0791212443
35 -583.0791212443
36 -583.0791212443
37 -583.0791212443
38 -583.0791212443
39 -583.0791212443
40 -583.0791212443
41 -583.0791212443
42 -583.0791212443
43 -583.0791212443
44 -583.0791212443
45 -583.0791212443
46 -583.0791212443
47 -583.0791212443
48 -583.0791212443
49 -583.0791212443
50 -583.0791212443
51 -583.0791212443
52 -583.0791212443
53 -583.0791212443
54 -583.0791212443
55 -583.0791212443
56 -583.0791212443
57 -583.0791212443
58 -566.1120722453
59 -566.1120722453
60 -566.1120722453
61 -566.1120722453
62 -566.1120722453
63 -566.1120722453
64 -566.1120722453
65 -566.1120722453
66 -566.1120722453
67 -566.1120722453
68 -566.1120722453
69 -566.1120722453
70 -566.1120722453
71 -566.1120722453
72 -566.1120722453
73 -566.1120722453
74 -566.1120722453
75 -566.1120722453
76 -566.1120722453
77 -566.1120722453
78 -566.1120722453
79 -566.1120722453
80 -566.1120722453
81 -566.1120722453
82 -566.1120722453
83 -566.1120722453
84 -566.1120722453
85 -566.1120722453
86 -566.1120722453
87 -566.1120722453
88 -566.1120722453
89 -566.1120722453
90 -566.1120722453
91 -566.1120722453
92 -566.1120722453
93 -566.1120722453
94 -566.1120722453
95 -566.1120722453
96 -566.1120722453
97 -566.1120722453
98 -566.1120722453
99 -566.1120722453
100 -566.1120722453
};
\addlegendentry{GE}
\addplot [semithick, color1]
table {%
0 -731.2375488742
1 -505.4781706265
2 -1407.392227513
3 -1322.191707022
4 -633.1751306426
5 -1681.635770624
6 -494.9641494678
7 -502.979313409
8 -474.8550990398
9 -468.3148503159
10 -462.2081618562
11 -453.1533026636
12 -442.0347198069
13 -436.0436798783
14 -444.0486088339
15 -430.6278533866
16 -479.9665250366
17 -521.3345398421
18 -507.0897248733
19 -1804.896203064
20 -483.6237856248
21 -487.5143944195
22 -503.930088544
23 -465.7452935654
24 -467.3134962708
25 -445.9277799437
26 -445.3996343303
27 -452.1006179612
28 -442.7727441176
29 -424.6504451313
30 -437.4110123676
31 -442.2404464325
32 -431.4999346594
33 -434.9596703406
34 -422.7717920763
35 -419.2454981469
36 -422.1185971327
37 -397.6908068083
38 -431.3708860735
39 -425.3013990073
40 -414.2662439599
41 -416.0519571766
42 -416.7592930848
43 -426.9647159782
44 -434.9457358907
45 -431.5902839708
46 -404.8860930747
47 -387.0471057306
48 -386.6543559957
49 -361.0257062643
50 -340.670207314
51 -322.7294143099
52 -301.7428503955
53 -309.0517408858
54 -312.2594202476
55 -307.7734918705
56 -305.725897745
57 -292.5539716018
58 -295.8656623781
59 -280.8783185538
60 -281.2412862792
61 -281.2524815789
62 -274.9779529866
63 -269.1345670092
64 -271.5049488747
65 -270.7330820192
66 -267.6215621854
67 -271.6335452237
68 -266.8542965188
69 -269.9714238462
70 -273.2615944973
71 -269.4285422179
72 -265.5355674276
73 -273.1804228888
74 -270.1504901654
75 -271.289915275
76 -271.2861981798
77 -268.0452392391
78 -273.1417715853
79 -271.9656946452
80 -277.9099080605
81 -272.5984060161
82 -275.8861215393
83 -272.2167347878
84 -272.8572913753
85 -274.5230890411
86 -271.2811872423
87 -268.94069954
88 -273.2678040144
89 -272.464848966
90 -266.204270271
91 -264.8662281133
92 -270.363924634
93 -267.3205630013
94 -268.6246205747
95 -268.646590498
96 -267.1908393191
97 -268.0832960603
98 -267.4373012049
99 -270.8992623337
100 -271.50695544
};
\addlegendentry{DRL}
\addplot [semithick, color2]
table {%
0 -354.2768905757
1 -354.2768905757
2 -353.4331412429
3 -353.4331412429
4 -339.5284883111
5 -339.5284883111
6 -339.5284883111
7 -322.319910118
8 -322.319910118
9 -322.319910118
10 -322.319910118
11 -322.319910118
12 -322.319910118
13 -322.319910118
14 -322.319910118
15 -322.319910118
16 -322.319910118
17 -322.319910118
18 -316.2552119412
19 -316.2552119412
20 -316.2552119412
21 -316.2552119412
22 -316.2552119412
23 -307.7651221113
24 -307.7651221113
25 -307.7651221113
26 -307.7651221113
27 -307.7651221113
28 -307.7651221113
29 -307.7651221113
30 -307.7651221113
31 -307.7651221113
32 -307.7651221113
33 -307.7651221113
34 -307.7651221113
35 -307.7651221113
36 -307.7651221113
37 -307.7651221113
38 -307.7651221113
39 -307.7651221113
40 -307.7651221113
41 -307.7651221113
42 -307.7651221113
43 -307.7651221113
44 -307.7651221113
45 -307.7651221113
46 -307.7651221113
47 -307.7651221113
48 -307.7651221113
49 -307.7651221113
50 -307.7651221113
51 -307.7651221113
52 -307.7651221113
53 -307.7651221113
54 -307.7651221113
55 -307.7651221113
56 -307.7651221113
57 -307.7651221113
58 -307.7651221113
59 -307.7651221113
60 -307.7651221113
61 -307.7651221113
62 -307.7651221113
63 -307.7651221113
64 -307.7651221113
65 -307.7651221113
66 -307.7651221113
67 -307.7651221113
68 -307.7651221113
69 -307.7651221113
70 -307.7651221113
71 -307.7651221113
72 -307.7651221113
73 -307.7651221113
74 -307.7651221113
75 -307.7651221113
76 -307.7651221113
77 -307.7651221113
78 -307.7651221113
79 -307.7651221113
80 -307.7651221113
81 -307.7651221113
82 -307.7651221113
83 -307.7651221113
84 -307.7651221113
85 -307.7651221113
86 -307.7651221113
87 -307.7651221113
88 -307.7651221113
89 -307.7651221113
90 -307.7651221113
91 -307.7651221113
92 -307.7651221113
93 -307.7651221113
94 -307.7651221113
95 -307.7651221113
96 -307.7651221113
97 -307.7651221113
98 -307.7651221113
99 -307.7651221113
100 -307.7651221113
};
\addlegendentry{MCTS}
\end{axis}

\end{tikzpicture}}
    \caption{The likelihood of the most likely failure found at each iteration of the GE, DRL, and MCTS solvers on the easy scenario, as well as GE+BA, DRL+BA, and MCTS+BA, which indicate in dashed lines the respective scores after robustification of each solver. Results are cropped to only show results when a failure was found.}
	\label{fig:results_easy}
	\vspace{-5mm}
\end{figure}
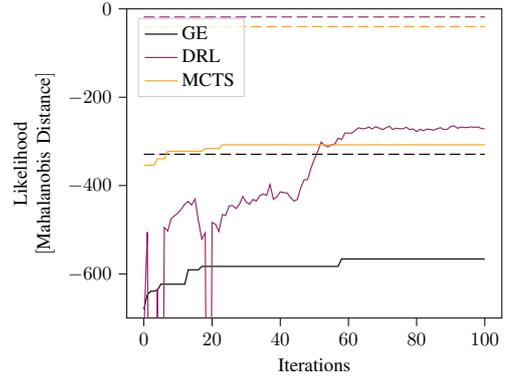

The results of all three solvers on the easy scenario are shown in \cref{fig:results_easy}.
All three algorithms are able to find failures quickly when given access to a heursitic reward. 
GE performs the worst, and both GE and MCTS show little improvement after finding their first failure. 
The DRL solver, however, continues to improve over the 100 iterations, ending with the best reward, and therefore the likeliest failure.
This scenario is contrived such that the likeliest actions lead to collision, and even the DRL solver still was not near the optimal reward of 0.
In contrast, adding robustification through BA resulted in significantly closer to optimal behavior.
While GE significantly improved with robustification, GE+BA was still outperformed by DRL.
However, both MCTS+BA and DRL+BA were able converge to results very near to 0.

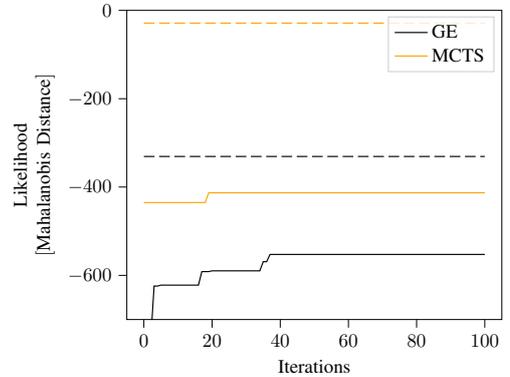
\begin{figure}[t]
	\centering
    \centering
    \scalebox{0.72}{
\begin{tikzpicture}

\definecolor{color0}{rgb}{0.001462,0.000466,0.013866}
\definecolor{color1}{rgb}{0.987622,0.64532,0.039886}

\begin{axis}[
font=\normalsize,
legend cell align={left},
legend style={fill opacity=0.8, draw opacity=1, text opacity=1, draw=white!80!black},
tick align=outside,
tick pos=left,
x grid style={white!69.0196078431373!black},
xlabel={Iterations},
xmin=-5, xmax=105,
xtick style={color=black},
y grid style={white!69.0196078431373!black},
ylabel style={align=center},
ylabel={Likelihood \\ $[$Mahalanobis Distance$]$},
ymin=-700, ymax=0,
ytick style={color=black}
]
\path [draw=color0, semithick, dash pattern=on 5.55pt off 2.4pt]
(axis cs:0,-330.7810346084)
--(axis cs:100,-330.7810346084);

\path [draw=color1, semithick, dash pattern=on 5.55pt off 2.4pt]
(axis cs:0,-28.813427422)
--(axis cs:100,-28.813427422);

\addplot [semithick, color0]
table {%
0 -759.1611322718
1 -747.7541654158
2 -747.7541654158
3 -624.2075695226
4 -624.2075695226
5 -622.0968500234
6 -622.0968500234
7 -622.0968500234
8 -622.0968500234
9 -622.0968500234
10 -622.0968500234
11 -622.0968500234
12 -622.0968500234
13 -622.0968500234
14 -622.0968500234
15 -622.0968500234
16 -622.0968500234
17 -591.4927704162
18 -591.4927704162
19 -591.4927704162
20 -589.7472741705
21 -589.7472741705
22 -589.7472741705
23 -589.7472741705
24 -589.7472741705
25 -589.7472741705
26 -589.7472741705
27 -589.7472741705
28 -589.7472741705
29 -589.7472741705
30 -589.7472741705
31 -589.7472741705
32 -589.7472741705
33 -589.7472741705
34 -589.7472741705
35 -569.0831501477
36 -569.0831501477
37 -552.6034805985
38 -552.6034805985
39 -552.6034805985
40 -552.6034805985
41 -552.6034805985
42 -552.6034805985
43 -552.6034805985
44 -552.6034805985
45 -552.6034805985
46 -552.6034805985
47 -552.6034805985
48 -552.6034805985
49 -552.6034805985
50 -552.6034805985
51 -552.6034805985
52 -552.6034805985
53 -552.6034805985
54 -552.6034805985
55 -552.6034805985
56 -552.6034805985
57 -552.6034805985
58 -552.6034805985
59 -552.6034805985
60 -552.6034805985
61 -552.6034805985
62 -552.6034805985
63 -552.6034805985
64 -552.6034805985
65 -552.6034805985
66 -552.6034805985
67 -552.6034805985
68 -552.6034805985
69 -552.6034805985
70 -552.6034805985
71 -552.6034805985
72 -552.6034805985
73 -552.6034805985
74 -552.6034805985
75 -552.6034805985
76 -552.6034805985
77 -552.6034805985
78 -552.6034805985
79 -552.6034805985
80 -552.6034805985
81 -552.6034805985
82 -552.6034805985
83 -552.6034805985
84 -552.6034805985
85 -552.6034805985
86 -552.6034805985
87 -552.6034805985
88 -552.6034805985
89 -552.6034805985
90 -552.6034805985
91 -552.6034805985
92 -552.6034805985
93 -552.6034805985
94 -552.6034805985
95 -552.6034805985
96 -552.6034805985
97 -552.6034805985
98 -552.6034805985
99 -552.6034805985
100 -552.6034805985
};
\addlegendentry{GE}
\addplot [semithick, color1]
table {%
0 -435.1187025794 
1 -435.1187025794
2 -435.1187025794
3 -435.1187025794
4 -435.1187025794
5 -435.1187025794
6 -435.1187025794
7 -435.1187025794
8 -435.1187025794
9 -435.1187025794
10 -435.1187025794
11 -435.1187025794
12 -435.1187025794
13 -435.1187025794
14 -435.1187025794
15 -435.1187025794
16 -435.1187025794
17 -435.1187025794
18 -435.1187025794
19 -413.0563966227
20 -413.0563966227
21 -413.0563966227
22 -413.0563966227
23 -413.0563966227
24 -413.0563966227
25 -413.0563966227
26 -413.0563966227
27 -413.0563966227
28 -413.0563966227
29 -413.0563966227
30 -413.0563966227
31 -413.0563966227
32 -413.0563966227
33 -413.0563966227
34 -413.0563966227
35 -413.0563966227
36 -413.0563966227
37 -413.0563966227
38 -413.0563966227
39 -413.0563966227
40 -413.0563966227
41 -413.0563966227
42 -413.0563966227
43 -413.0563966227
44 -413.0563966227
45 -413.0563966227
46 -413.0563966227
47 -413.0563966227
48 -413.0563966227
49 -413.0563966227
50 -413.0563966227
51 -413.0563966227
52 -413.0563966227
53 -413.0563966227
54 -413.0563966227
55 -413.0563966227
56 -413.0563966227
57 -413.0563966227
58 -413.0563966227
59 -413.0563966227
60 -413.0563966227
61 -413.0563966227
62 -413.0563966227
63 -413.0563966227
64 -413.0563966227
65 -413.0563966227
66 -413.0563966227
67 -413.0563966227
68 -413.0563966227
69 -413.0563966227
70 -413.0563966227
71 -413.0563966227
72 -413.0563966227
73 -413.0563966227
74 -413.0563966227
75 -413.0563966227
76 -413.0563966227
77 -413.0563966227
78 -413.0563966227
79 -413.0563966227
80 -413.0563966227
81 -413.0563966227
82 -413.0563966227
83 -413.0563966227
84 -413.0563966227
85 -413.0563966227
86 -413.0563966227
87 -413.0563966227
88 -413.0563966227
89 -413.0563966227
90 -413.0563966227
91 -413.0563966227
92 -413.0563966227
93 -413.0563966227
94 -413.0563966227
95 -413.0563966227
96 -413.0563966227
97 -413.0563966227
98 -413.0563966227
99 -413.0563966227
100 -413.0563966227
};
\addlegendentry{MCTS}
\end{axis}

\end{tikzpicture}}
    \caption{The likelihood of the most likely failure found at each iteration of the GE and MCTS solvers on the medium scenario, as well as GE+BA and MCTS+BA, which indicate in dashed lines the respective scores after robustification of each solver. The DRL solver was unable to find a failure. Results are cropped to only show results when a failure was found.}
	\label{fig:results_medium} 
	\vspace{-5mm}
\end{figure}

\Cref{fig:results_medium} shows the results of the non-DRL solvers on the medium difficulty scenario. 
Without a heuristic to provide reward signal, the DRL solver was unable to find a failure within 100 iterations.
With the short horizon, MCTS is still able to outperform GE. 
Adding a robustification phase again improves both algorithms, and again the MCTS+BA solver outperforms the GE+BA solver.
Note that taking the average action is not sufficient to cause a crash in this scenario.

\begin{figure}[t]
	\centering
    \centering
    \scalebox{0.72}{
\begin{tikzpicture}

\definecolor{color0}{rgb}{0.001462,0.000466,0.013866}

\begin{axis}[
font=\normalsize,
legend cell align={left},
legend style={fill opacity=0.8, draw opacity=1, text opacity=1, draw=white!80!black},
tick align=outside,
tick pos=left,
x grid style={white!69.0196078431373!black},
xlabel={Iterations},
xmin=-5, xmax=105,
xtick style={color=black},
y grid style={white!69.0196078431373!black},
ylabel style={align=center},
ylabel={Likelihood \\ $[$Mahalanobis Distance$]$},
ymin=-1800, ymax=-400,
ytick style={color=black}
]
\path [draw=color0, semithick, dash pattern=on 5.55pt off 2.4pt]
(axis cs:0,-610.9315394596)
--(axis cs:100,-610.9315394596);

\addplot [semithick, color0]
table {%
0 -1307.5534305427
1 -1284.1922241497
2 -1284.1922241497
3 -1284.1922241497
4 -1284.1922241497
5 -1284.1922241497
6 -1284.1922241497
7 -1284.1922241497
8 -1284.1922241497
9 -1284.1922241497
10 -1284.1922241497
11 -1284.1922241497
12 -1284.1922241497
13 -1281.2324817686
14 -1281.2324817686
15 -1281.2324817686
16 -1281.2324817686
17 -1281.2324817686
18 -1281.2324817686
19 -1281.2324817686
20 -1281.2324817686
21 -1281.2324817686
22 -1281.2324817686
23 -1272.3850555586
24 -1272.3850555586
25 -1267.6505634676
26 -1267.6505634676
27 -1213.1255412785
28 -1213.1255412785
29 -1213.1255412785
30 -1213.1255412785
31 -1213.1255412785
32 -1213.1255412785
33 -1213.1255412785
34 -1213.1255412785
35 -1213.1255412785
36 -1199.259524253
37 -1196.1941678764
38 -1196.1941678764
39 -1196.1941678764
40 -1196.1941678764
41 -1196.1941678764
42 -1196.1941678764
43 -1196.1941678764
44 -1196.1941678764
45 -1196.1941678764
46 -1196.1941678764
47 -1196.1941678764
48 -1196.1941678764
49 -1196.1941678764
50 -1196.1941678764
51 -1196.1941678764
52 -1196.1941678764
53 -1196.1941678764
54 -1196.1941678764
55 -1196.1941678764
56 -1196.1941678764
57 -1196.1941678764
58 -1196.1941678764
59 -1196.1941678764
60 -1196.1941678764
61 -1196.1941678764
62 -1196.1941678764
63 -1196.1941678764
64 -1196.1941678764
65 -1196.1941678764
66 -1196.1941678764
67 -1196.1941678764
68 -1196.1941678764
69 -1196.1941678764
70 -1196.1941678764
71 -1196.1941678764
72 -1196.1941678764
73 -1196.1941678764
74 -1196.1941678764
75 -1196.1941678764
76 -1196.1941678764
77 -1196.1941678764
78 -1196.1941678764
79 -1196.1941678764
80 -1196.1941678764
81 -1196.1941678764
82 -1196.1941678764
83 -1196.1941678764
84 -1196.1941678764
85 -1196.1941678764
86 -1196.1941678764
87 -1196.1941678764
88 -1196.1941678764
89 -1196.1941678764
90 -1196.1941678764
91 -1196.1941678764
92 -1196.1941678764
93 -1196.1941678764
94 -1196.1941678764
95 -1196.1941678764
96 -1196.1941678764
97 -1196.1941678764
98 -1196.1941678764
99 -1196.1941678764
100 -1140.7594249208
};
\addlegendentry{GE}
\end{axis}

\end{tikzpicture}}
    \caption{The likelihood of the most likely failure found at each iteration of the GE solver on the hard scenario, as well as GE+BA, which indicates in a dashed line the score after robustification. The DRL  and MCTS solvers were unable to find a failure.}
	\label{fig:results_hard} 
	\vspace{-5mm}
\end{figure}
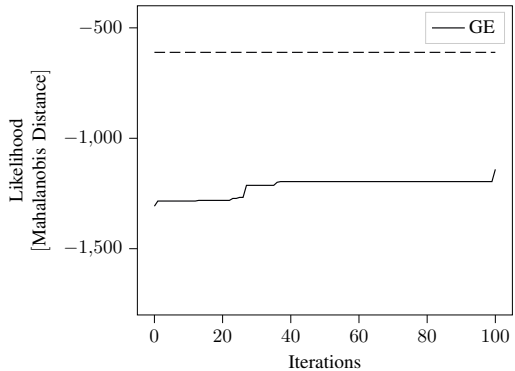

\Cref{fig:results_hard} shows the results of GE and GE+BA on the hard difficulty scenario.
The hard scenario has a longer horizon, which prevented both DRL and MCTS from being able to find failures within 100 iterations. 
GE was still able to find failures, and GE+BA was still able to improve the results.
In fact, when adjusting for the increased number of steps, the GE and GE+BA results on the hard scenario are very similar to those on the medium scenario, showing that GE is robust in longer horizon problems.

The results across the three scenarios illuminate the strengths and weaknesses of the three algorithms. 
When a useful reward signal is present, the DRL solver shows the best ability to find the most likely failure.
However, without a heuristic, it quickly loses its ability to find failures. 
In a no-heuristic setting, MCTS is able to find more likely failures than GE as long as the problem's horizon is short. 
However, in longer horizon problems GE is able to find failures that MCTS cannot
The underlying principle of these differences is how the algorithms balance exploration and exploitation, which also explains why adding a robustification phase is consistently able to improve results across all three algorithms.
The robustification phase applies the strength of DRL, exploitation, to a domain that requires significantly less exploration.
Consequently, these two-phase methods could also be seen as an exploration phase and an exploitation phase, a problem decomposition that is easier to solve.

\section{Conclusion}

This paper presented an approach for validating autonomous vehicles without the use of heuristic rewards.
Without a heuristic reward, the RL agent does not have a reward signal guiding it towards its goal, a type of problem known as hard-exploration .
We used a modified version of go-explore, a state-of-the-art algorithm for hard-exploration.
GE was able to find failures on longer horizon problems where MCTS and DRL could not.
Furthermore, inspired by the robustification phase of GE, we were able to use the backwards algorithm to improve the results of all three solvers.
Future work can focus further on decomposing validation into an exploration phase and an exploitation phase to take advantage of the strengths of different solvers.
\bibliographystyle{IEEEtran}
\bibliography{AST2}

\end{document}